\def\year{2021}\relax
\documentclass[letterpaper]{article} 
\usepackage{aaai21}  
\usepackage{times}  
\usepackage{helvet} 
\usepackage{courier}  
\usepackage[hyphens]{url}  
\usepackage{graphicx} 
\usepackage{graphicx}  
\usepackage{natbib}  
\usepackage{caption} 
\usepackage{CJK}  
\usepackage{color}
\usepackage{setspace}  
\usepackage{bm}
\usepackage{amsmath}
\usepackage[linesnumbered,ruled,lined]{algorithm2e}
\usepackage[switch]{lineno}  %

\title{SARG: A Novel Semi Autoregressive Generator for Multi-turn Incomplete Utterance Restoration}

\author{Mengzuo Huang$^{1,2}$\thanks{\quad Equal contribution. This work was conducted when Mengzuo Huang was interning at Netease Games AI Lab} \quad Feng Li$^{2*}$ \quad Wuhe Zou$^2$ \quad Weidong Zhang$^{2}$\thanks{\quad Corresponding author} \\
}

\affiliations{
	\textsuperscript{\rm 1} Dalian University of Technology \\
	\textsuperscript{\rm 2} Netease Games AI Lab, HangZhou, China
	
	hmengzuo@mail.dlut.edu.cn \\
	\{lifeng06,zouwuhe,zhangweidong02\}@corp.netease.com
}

\begin{document}
\maketitle

\begin{abstract}

Dialogue systems in open domain have achieved great success due to the easily obtained single-turn corpus and the development of deep learning, but the multi-turn scenario is still a challenge because of the frequent coreference and information omission. 
In this paper, we investigate the incomplete utterance restoration which has brought general improvement over multi-turn dialogue systems in recent studies. 
Meanwhile, jointly inspired by the autoregression for text generation and the sequence labeling for text editing, we propose a novel semi autoregressive generator (SARG) with the high efficiency and flexibility. 
\textcolor{black}{Moreover, experiments on two benchmarks  show that our proposed model significantly outperforms the state-of-the-art models in terms of quality and inference speed.} \footnote{https://github.com/NetEase-GameAI/SARG}	         
\end{abstract}                                          

\section{Introduction}


Dialogue systems in open-domain have attracted increasing attention \cite{li2020empirical, huang2020challenges}, and been widely utilized in real-world applications  \cite{adiwardana2020towards, gong2019customer, hewitt2020case}.
However, due to frequently occurred coreference and information omission, as shown in Table \ref{table_example}, there still exists a major challenge: it is hard for machines to understand the real intention from the original utterance without the context.
A series of models of retrieval-based and generative-based have been studied for multi-turn systems \cite{yan2016learning, zhang2018modeling, zhou2018multi, wu2016sequential}, and they generally combine the context and the original utterance as input to retrieve or generate responses. 
However, these methods lack great generalizations since they have a strong reliance on the size of the multi-turn corpus. 

\begin{CJK}{UTF8}{gbsn}
	\renewcommand{\arraystretch}{1.2}
	\newcommand{\hanzi}{\fontsize{8.5pt}{\baselineskip}\selectfont}
	\begin{table}[h]
		\small
		\begin{center}
			\begin{tabular}{l|p{5cm}}
				\hline
				& \textbf{Context 1}\\
				Utterance 1 & Human: \hanzi{为什么？}\\
				(Translation)	& Human:\textit{ Why?} \\ \hline
				Utterance 2 & Chatbot: \hanzi{这个你得问\textcolor{red}{李淳风}呀。}  \\
				& Chatbot: \textit{You'll have to ask \textcolor{red}{Li Chunfeng} about that.} \\ \hline
				Utterance 3 & Human: \hanzi{我去问\textcolor{red}{他}。} \\ 
				& Human: \textit{I'll ask \textcolor{red}{him}.} \\ \hline
				Utterance $3^{\prime}$  & Human: \hanzi{我去问\textcolor{red}{李淳风}。}  \\
				& Human: \textit{I'll ask \textcolor{red}{Li Chunfeng}.} \\ \hline
				& \textbf{Context 2}\\
				Utterance 1 & Human: \hanzi{你\textcolor{blue}{最喜欢}什么电影？}\\
				& Human:\textit{What movie do you \textcolor{blue}{like most}?} \\ \hline
				Utterance 2 & Chatbot: \hanzi{\textcolor{blue}{泰坦尼克}。}  \\
				& Chatbot: \textit{\textcolor{blue}{Titanic}.} \\ \hline
				Utterance 3 & Human: \hanzi{为什么呢？} \\ 
				& Human: \textit{Why?} \\ \hline
				Utterance $3^{\prime}$  & Human: \hanzi{为什么\textcolor{blue}{最喜欢泰坦尼克}？}  \\
				& Human: \textit{Why do you \textcolor{blue}{like Titanic most}?} \\ \hline
				
			\end{tabular}
		\end{center}
		\caption{An example of utterance restoration in human-machine dialogue system. Utterance $3^{\prime}$ is the restored sentence based on Utterance 3. Red means
			coreference and blue means omission.}\label{table_example}
	\end{table}
\end{CJK}

\citealt{su2019improving} and \citealt{pan2019improving} propose their utterance restoration models, respectively, which are aimed at restoring the semantic information of the original utterance based on the history of the session from a different perspective.
Restoration methods decouple multi-turn systems into the single-turn problems, which alleviate the dependence on multi-turn dialogue corpus and also achieve leading performance. 
Specifically, \citealt{su2019improving} employ transformer-based Seq2Seq architecture and pointer network to rewrite the original utterance, and they split the whole session into history and original utterance for capturing different attentions.
\citealt{pan2019improving} propose a cascade frame of ``pick-and-combine" to restore the incomplete utterance from history. 
And both of them generate restored utterance from scratch in an autoregressive manner of  Seq2Seq, which is highly time-consuming during inference.

Unlike some traditional end-to-end text generation task, where the apparent disparity exists between the sources and targets, utterance restoration always has some considerable overlapping regions between inputs and outputs. 
Intuitively, some sequence labeling methods can be utilized to speed up the inference stage in this task, since Seq2Seq from scratch is time wasteful.
Further, \citealt{malmi2019encode} introduce LaserTagger, a sequence labeling method, which casts text generation as a text editing task. 
However, the insertions of LaserTagger are restricted to a fixed phrase vocabulary that is derived from the training data.
In multi-turn dialogue, some rare phrases are habitually omitted by the speaker without affecting the listening comprehension; as shown in Table \ref{table_example}, ``\textit{Li Chunfeng}" is a rare phrase and omitted in Utterance 3 of Context 1. And LaserTagger can not solve such a coreference problem well, since the rare phrase is discarded when constructing the fixed phrase vocabulary.

As a first attempt to combine the sequence labeling and autoregression in utterance restoration, we propose a semi autoregressive generator (SARG), which can well tackle the challenges brought by highly time-consuming and discarded rare words or phrases.
SARG retains the flexibility of autoregression and takes advantage of the fast inference speed of sequence labeling.

First, we employ a tagger to predict the editing labels, which involves three main operations: \texttt{KEEP} a token, \texttt{DELETE} a token, \texttt{CHANGE} a token with other phrases.
Then, instead of adding phrases from a pre-defined phrase vocabulary, we utilize an autoregressive decoder based on LSTM with copy mechanism for generating the added phrases.
Moreover, inspired by the great success of the pretrained transformer models \cite{vaswani2017attention}, we also design an encoder based on BERT \cite{devlin2018bert} to obtain the contextual encodings. 
\textcolor{black}{Finally, we perform experiments on two benchmarks: the \textit{Restoration-200k} \cite{pan2019improving} and \textit{CANARD} \cite{elgohary-etal-2019-unpack}, the SARG shows superiorities on the automatic evaluation, the human evaluation, and the inference speed respectively.}
In summary, our contributions are:

\begin{itemize}
	\setlength{\itemsep}{2pt}
	\setlength{\parsep}{2pt}
	\setlength{\parskip}{2pt}
	\item SARG is a creative fusion of sequence labeling and autoregressive generation, which is suitble for utterance restoration task;
	\item SARG solves the restoration problem by a joint way and can easily load the pretrained BERT weights for the overall model;
	\item SARG obtains a competitive performance and faster inference speed.
\end{itemize}


\section{Related Work}
\label{related}
\subsection{Multi-turn Dialogue systems}

Recently, building a chatbot with data-driven approaches in open-domain has drawn significant attention \cite{ritter2011data,ji2014information,athreya2018enhancing}.
Most of works on conversational systems can be divided into retrieval-based methods \cite{ji2014information,yan2016learning,zhou2016multi,wu2016ranking,wu2016sequential,zhou2018multi,zhang2018modeling} and generation-based methods \cite{serban2016building,xing2016topic,serban2017multiresolution,zhao2020learning,lin2020hierarchical}. Though the above methods are enlightening, there is a lack of high-quality multi-turn dialogue data to train them. 


In multi-turn dialogue systems, existing methods are still far from satisfactory compared to the single-turn ones, since
the coreference
and information omission frequently occur in our daily conversation, which  
makes machines hard to understand
the real intention \cite{su2019improving}.
Recent studies suggest simplifying the multi-turn dialogue modeling
into a single-turn problem by restoring the incomplete
utterance \cite{su2019improving, pan2019improving}. 
\citealt{su2019improving} rewrite the utterance based on transformer-based Seq2Seq and pointer network from context with two-channel attentions. 
\citealt{pan2019improving} propose a cascaded ``pick-and-combine"
model to restore the incomplete utterance
from its context.
Moreover, \citealt{pan2019improving} release the high quality datasets \textit{Restoration-200k} for the study of incomplete utterance restoration in open-domain dialogue systems.

\subsection{Sentence Rewriting}
Sentence rewriting is a general task which has
high overlap between input text
and output text,
such as: 
text summarization\cite{see2017get,chen2018fast,cao2018retrieve}, text simplification\cite{wubben2012sentence,zhang2017sentence}, grammatical error correction\cite{ng2014conll,ge2018fluency,chollampatt2018neural,zhao2019improving} and sentence fusion \cite{thadani2013supervised,lebanoff2019analyzing}, ect.
Seq2Seq model, which provides a powerful framework for learning to translate
source texts into target texts, is the main approach for sentence rewriting.
However, conventional Seq2Seq approaches require large amounts of training data and take low-efficiency on inference.  

\citealt{malmi2019encode} propose a sequence labeling approach for sentence rewriting that casts text generation as a text editing task. 
And the method is fast enough at inference time with performance
comparable to the state-of-the-art Seq2Seq models.
However, it can't be applied to our incomplete utterance restoration well, due to some limitations of inflexibility.

To make full use of the flexibility of autoregressive models and the efficiency of sequence labeling models, we combine the autoregressive generation and the sequence labeling for the trade-off between inference time and model flexibility.

\section{Methodology}
\label{method}
\begin{figure*}[htb]
	\centering
	\includegraphics[width=1.0\linewidth]{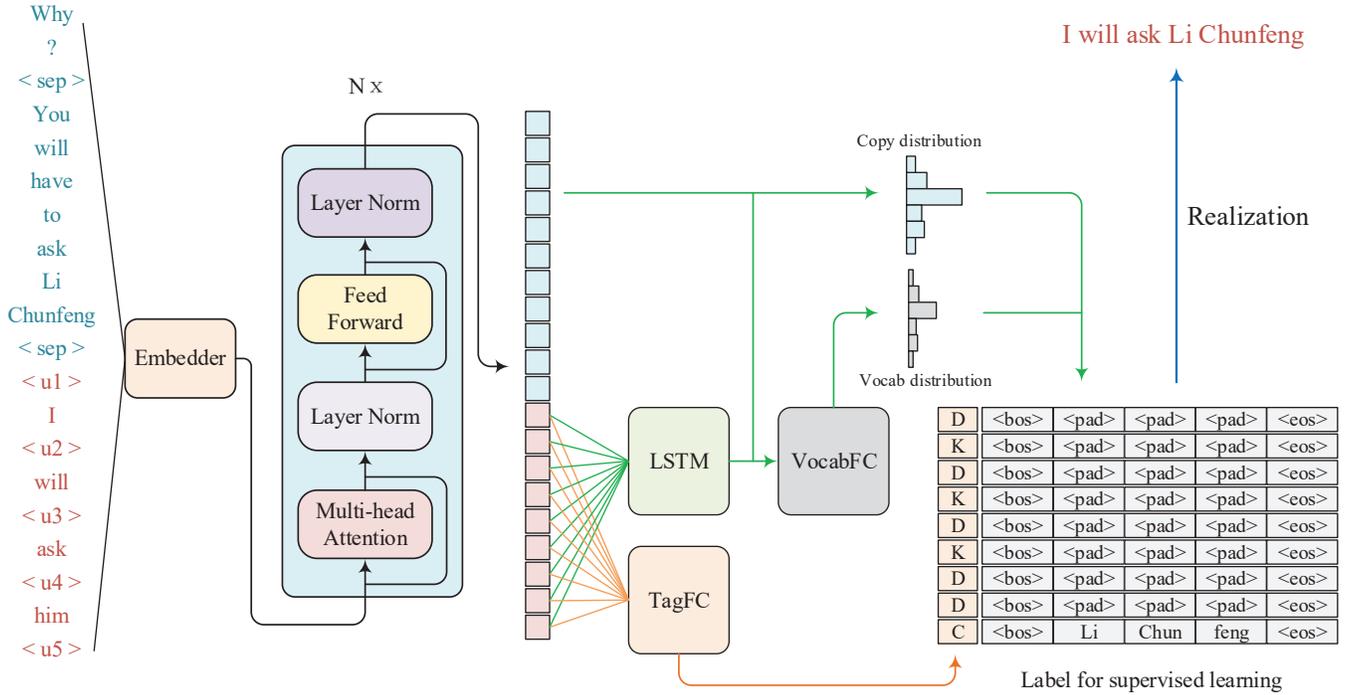}
	\caption{The overall architecture of the proposed SARG. In the constructed label, \textbf{D} means the \texttt{DELETE} operation, \textbf{K} means the \texttt{KEEP}, \textbf{C} means the \texttt{CHANGE} and the  phrase of \textit{``Li Chunfeng''} is the added phrase for this \texttt{CHANGE} operation. In the input, the blue words are the history of the session, the red words are the original utterance and the $<$ui$>$ is the dummy token. In the dataflow, the black means encoding, the orange means tagging, the green means decoding, and the blue means the realization.}\label{fig:sarg}
\end{figure*}

In this section, we demonstrate our proposed SARG for the multi-turn incomplete utterance restoration. 
The restoration problem can be denoted as $f(H, U) = R$, where $H = \{\bm{w}_1^h, \bm{w}_2^h, ..., \bm{w}_m^h\}$ is the history of dialogue (context), 
$U = \{\bm{w}_1^u, \bm{w}_2^u, ..., \bm{w}_n^u\}$ is the original utterance (source) to be rewritten and $R$ is the restored utterance (target). 
The overall architecture of SARG is shown in Figure \ref{fig:sarg}. Instead of generating the restored utterance from scratch as traditional Seq2Seq, we first determine the editing operation sequence across the original utterance; then generate the potential phrases according to the operation sequence; finally convert the operation sequence and the generated phrases to text.
The detailed descriptions are as follows.\footnote{By convention, the bold letters represent the vectors, the capital letters represent the matrices and others represent the scalars. }

\subsection{Tagging Operations}

\begin{algorithm}[!p]
	\caption{Convert the target to label}\label{alg1}
	\LinesNumbered
	\setstretch{0.88}
	
	\KwIn{$S$: the original utterance \newline $T$: the restored utterance}
	\KwOut{$L$: the supervised label}
	
	Insert dummy tokens in $S$
	
	$L[i] = \texttt{DELETE}, \forall i = 1, 2, ..., 2n+1$
	
	$j = 0$; $k = 0$; $\texttt{A} = [\ ]$
	
	Compute the \textit{longest common subsequence} $K$ between $S$ and $T$
	
	\For{$i \in [1, 2n+1]$}{
		
		\If{$S[i] = K[k]$}{
			$L[i] = \texttt{KEEP}$ \\
			\While{$T[j] \neq K[k]$}{
				
				$\texttt{A} = \texttt{A} + T[j]$ \\
				$j = j + 1$
				
			}
			
			$k = k + 1$
			
			\If{$\texttt{A} \neq \emptyset$}{
				$L[i - 1] = \texttt{CHANGE}\ \texttt{A}$\\
				$\texttt{A} = [\ ]$
			}
		}
	}
	\If{$T[j:] \neq \emptyset$}{
		$\texttt{A} = T[j:]$ \\
		$L[-1] = \texttt{CHANGE}\ \texttt{A}$
	}
	
	\Return $L$
\end{algorithm}

First of all, meaningless dummy tokens are inserted between every two tokens in the original utterance, as shown in first column of Figure \ref{fig:sarg}. We can directly add the phrases in the gaps between every two tokens by the insertion of dummy tokens, which eliminates the ambiguity of possible editing operations to some extent.
Moreover, we recommend that the original tokens can only be kept or deleted, and the dummy tokens can only be deleted or changed by other phrases. 

Formally, three editing operations are defined in this work: \texttt{KEEP}, \texttt{DELETE} and \texttt{CHANGE}. 
Intuitively, \texttt{KEEP} means that the token remains in the restored utterance, \texttt{DELETE} means that the token is undesired, and \texttt{CHANGE} $\texttt{A}$ means that the token should be replaced by the informative phrase $\texttt{A}$.

\begin{table}[!h]
	\centering
	\scalebox{0.95}{
		\begin{tabular}{lcc}
			\hline
			& Added Phrase & Restored Utterance \\ \hline
			Avg. length & 3.1 & 12.4 \\ \hline
		\end{tabular}
	}
	\caption{Comparison of the average length between the added phrase and the restored utterance on \textit{Restoration-200k}.}\label{comparison}
\end{table}

The following steps are employed to construct the supervised labels: (1) first compute the \textit{longest common subsequence} (LCS) between original and restored utterance; (2) then greedily attempt to align the original utterance, restored utterance and the LCS; (3) finally replace the undesired tokens in original utterance with the added tokens in restored utterance.
The detailed descriptions are demonstrated in Algorithm \ref{alg1}, and the constructed labels can be referenced in Figure \ref{fig:sarg}. Specifically, the first column of labels is used to supervise the tagger and other columns are used for the decoder.

Moreover, the comparison of average length between added phrase and the restored utterance is listed in Table \ref{comparison}, which indicates that SARG saves at least three-quarters of the time for decoding compared to those complete autoregressive model.

\subsection{Encoder}
Since pretrained transformers \cite{vaswani2017attention} have been shown to be beneficial in many downstream NLP tasks \cite{radford2018improving, devlin2018bert}, in this work, we utilize the standard transformer blocks as the backbone of the encoder, like the black lines in Figure 1.

In the embedding module, we concatenate the history $H$ and the original utterance $U$ (involved dummy tokens) as the input sequence $W = \{\bm{w}_1, \bm{w}_2, ..., \bm{w}_k\}$, then embed them into continuous space by looking up the following embedding tables:
\begin{itemize}
	\setlength{\itemsep}{3pt}
	\setlength{\parsep}{3pt}
	\setlength{\parskip}{3pt}
	\item Word Embedding: the word embedding table is built on a pre-defined wordpiece vocabulary from pretrained transformers.
	\item Position Embedding: the position embedding table is also initialized by pretrained transformers.
	\item Turn Embedding: turn embedding is used to indicate which turn each token belongs to. The looking-up table is randomly initialized.
\end{itemize}
For each token $\bm{w}_i$, we sum and normalize \cite{ba2016layer} the above three embeddings, then acquire the input embedding:
\begin{equation}\label{key}
E^{(0)}_i = \textrm{LN}(W\hspace{-1pt}E(\bm{w}_i) + P\hspace{-1pt}E(\bm{w}_i) + T\hspace{-1pt}E(\bm{w}_i)),
\end{equation}
where $W\hspace{-1pt}E$ is the word embedding, $P\hspace{-1pt}E$ is the position embedding and $T\hspace{-1pt}E$ is the turn embedding. Once the input embedding is acquired, we feed such representation into the $L$ stacked transformer blocks:
\begin{equation}\label{key}
E^{(l)} = \textrm{TransformerBlock}(E^{(l-1)}).
\end{equation}
At last, we obtain the final encodings $E^{(L)}$, which can be further divided into two parts according to the partitions of history and original utterance:
\begin{linenomath}
\begin{align}
	E_{h} &= \{\bm{h}_1, \bm{h}_2, \cdots, \bm{h}_m\}, \\
	E_{u} &= \{\bm{u}_1, \bm{u}_2, \cdots, \bm{u}_{2n+1}\},
\end{align}
\end{linenomath}
where $E_{h}$ is the encodings of history and $E_{u}$ is the encodings of original utterance. There are $n+1$ dummy tokens in the original utterance, which collect the information from those original tokens by the self-attention.

\begin{table*}[!htb]
	\centering
	
	\begin{tabular}{lccccccc}
		\hline
		Model & $f_1$ & $f_2$ & $f_3$ & BLEU-1 & BLEU-2 & ROUGE-1 & ROUGE-2  \\ \hline
		CopyNet & 50.3  & 41.1  & 34.9  & 84.7  & 81.7  & 89.0  & 80.9    \\ 
		T-Ptr-$\lambda$ & 51.0 & 40.4 & 33.3  & 90.3 & 87.4 & 90.1 & 83.0 \\ 
		PAC$^\ddag$ & \textbf{63.7} & 49.7 & 40.4 & 89.9 & 86.3 & 91.6 & 82.8   \\ 	
		Seq2Seq-Uni$^\ddag$  & 56.8 & 46.4 & 39.8  & 90.8 & 88.3 & 91.4 & 85.0  \\ 
		SARG$^\ddag$ & 62.4 & \textbf{52.5} & \textbf{46.3}  & \textbf{92.2} & \textbf{89.6} & \textbf{92.1} & \textbf{86.0} \\ 
		\hline
	\end{tabular}
	\caption{The main results on \textit{Restoration-200k} of our method and other SOTA methods. The models with ``$\ddag$" means that pretrained weights like BERT are utilized. Except to SARG, other models employ the 5-beam-search in their decoding procedure. SARG employs the greedy search in decoding step.} \label{200k}
\end{table*}

\subsection{Tagger}
Tagger takes the encodings $E_u$ as the input and predicts the editing labels on each token in original utterance. As shown in Figure \ref{fig:sarg}, the orange lines stand for the dataflow of tagger. In our setting, a single linear transformation layer with softmax activation function is employed for projecting the encoding to the space of editing labels, the formula is as follows:
\begin{equation}\label{key}
\bm{p}(\bm{y}_i|\bm{u}_i) = \textrm{softmax}(W_t \cdot \bm{u}_i+ b_t),
\end{equation}
where $W_{t}$ and $b_{t}$ are parameters to be learned, and the following $W$ and $b$ are all learnable. 
Finally, the loss provided by the tagger is defined as negative log-likelihood:
\begin{equation}\label{key}
\textrm{loss}_{tag} = - \sum_{i} \log \bm{p}(\bm{y}_i|\bm{u}_i),
\end{equation}
where $i$ is corresponding to the index of token in original utterance.
\subsection{Decoder}
Different from the general autoregressive decoder that performs decoding from scratch, in our setting, the decoder, as green lines in Figure 1, works in parallel on the tokens which get \texttt{CHANGE} operations in tagger. \textcolor{black}{Specifically, the decoder is only one and shared by these tokens.}

For the consideration of efficiency, we employ one layer of unidirectional LSTM \cite{hochreiter1997long} as the backbone of our decoder. 
For each token in original utterance, the related initial state $\bm{s}_0$ is initialized with the according hidden representation \footnote{It is a remarkable fact that there is a one-to-one correspondence between the hidden representation $\bm{u}_i$ and the state $\bm{s}_0$, however, we omit the subscript $i$ in $\bm{s}_0$ for the convenient expression.}:
\begin{equation}\label{key}
\bm{s}_0 = \bm{u}_i \in E_u.
\end{equation}
Then the autoregressive generation is described as follows:
\begin{equation}\label{key}
\bm{s}_t = \textrm{LSTM}(W\hspace{-2pt}E(\bm{x}_t), \bm{s}_{t-1}),
\end{equation}
where $\bm{x}_t$ is the output of decoder in the previous step, and the $\bm{x}_1$ is initialized by a special start token.

Moreover, in order to dynamically choose copying from the history or sampling from the overall vocabulary, we introduce the recurrent attention and coverage mechanism as in pointer-generator network \cite{see2017get}. At each decoding step, we utilize the output $\bm{s}_t$ to collect information from the encodings of history $E_h$. The detailed calculations are as follows:
\begin{linenomath}
\begin{align}
	\bm{e^{t}}_j &= \bm{v}^T \tanh(W_s \bm{s}_t + W_{h} \bm{h}_j + w_c \bm{c^{t}}_j + b_{attn}), \\
	\bm{a^{t}} &= \textrm{softmax}(\bm{e^{t}}),
\end{align}
\end{linenomath}
where $j$ is corresponding to the index of token in the history, $t$ is corresponding to the decoding steps and the $\bm{c^{t}}$ is the coverage vector in $t$-th step. Specifically, the coverage vector is initialized by zero at the beginning of decoding and accumulated as follow:
\begin{equation}\label{key}
\bm{c^{t}}_j = \sum_{t^{\prime}=0}^{t-1} \bm{a^{t^{\prime}}}_j.
\end{equation}
Once the normalized weights $\bm{a}^t$ are obtained, we can calculate the results of attention:
\begin{equation}\label{key}
\bm{s^*_{t}} = \sum_{j} \bm{a^{t}}_j \cdot \bm{h}_j.
\end{equation}
Then, the $\bm{s^*_{t}}$ is forwarded into the subsequent modules for acquiring the predicted word:
\begin{linenomath}
\begin{align}
	g &= \sigma (\bm{w_{s^*}}^T \bm{s^*_{t}} + \bm{w_{s}}^T \bm{s_t} + \bm{w_x}^T W\hspace{-2pt}E(\bm{x_t}) + b_g), \\
	\bm{p_{vocab}} &= \textrm{softmax}(W_v \cdot \bm{s^*}_t + b_v),~~~~~~~~ \\
	\hspace{-5pt}\bm{p}(\bm{x_{t+1}}) &= g \cdot \bm{p_{vocab}} + (1 - g) \sum_{j:w_j=x_t} \bm{a^t}_{j}, 
\end{align}
\end{linenomath}
where $\sigma$ is the sigmoid function
to output a value between $0$ and $1$, the $g$ is the gate to make a trade-off between copying and generating, the $\bm{p}(\bm{x_{t+1}})$ is the final probability distribution of generated word. Moreover, the coverage loss is introduced to penalize repeatedly attending:
\begin{equation}\label{key}
\textrm{covloss}_{t} = \sum_j \min (\bm{a^{t}}_j, \bm{c^{t}}_j).
\end{equation}
Finally, the loss of the decoder is the weighted sum of negative log-likelihood and the coverage loss:
\begin{equation}\label{key}
\textrm{loss}_{dec} = \sum_i \sum_t - \log \bm{p}(\bm{x_t}^i) + \lambda \ \textrm{covloss}^i_{t},
\end{equation}
where $i$ is corresponding to the index of token in original utterance, $\lambda$ is the hyperparameter for adjusting the weight.

\subsection{Joint Training}
The model is optimized jointly. Once the loss of tagger and decoder are obtained, we sum and backward propagate the total loss as below:
\begin{equation}\label{key}
\textrm{loss} = \alpha\ \textrm{loss}_{tag} + \textrm{loss}_{dec},
\end{equation}
where $\alpha$ is also the hyperparameter for adjusting the weight.

\subsection{Realization}
In the realization, we convert the predicted editing labels and the generated phrases to a complete utterance. In detail, we remain the \texttt{KEEP} denoted token and remove the \texttt{DELETE} token in the original utterance (involved dummy tokens), and replace the token, assigned by \texttt{CHANGE A}, with the generated phrase \texttt{A}.

\section{Experiments}
\label{exp}
\begin{table}[h]
	\centering
	\begin{tabular}{cccc}
		\hline
		& train & dev & test \\ \hline
		\textit{Restoration-200k} & 194k & 5k & 5k \\ 
		\textit{CANARD} & 32k &4k &6k \\
		\hline
	\end{tabular}
	\caption{The count of conversations in different datasets.}\label{rstat}
\end{table}

\begin{table*}[!htb]
	\centering
	\begin{tabular}{lccccccc}
		\hline
		& $f_1$ & $f_2$ & $f_3$  & BLEU-1 & BLEU-2 & ROUGE-1 & ROUGE-2\\ \hline
		SARG  & 62.4 & 52.5 & 46.3  & 92.2 & 89.6 & 92.1 & 86.0 \\
		\ w/o WEIGHT  & 52.8 & 41.1 & 33.8 & 89.2 & 86.7 & 89.9 & 83.6 \\
		\ w/o COPY  & 55.6 & 38.9 & 32.8 & 89.4 & 85.6 & 89.9 & 81.7 \\
		\ w/o GEN & 56.2 & 48.0 & 42.9 & 90.4 & 88.2 & 91.4 & 85.6 \\ \hline
	\end{tabular}
	\caption{Ablation study of proposed model on the \textit{Restoration-200K}. The beam size is fixed to 1.} \label{ablation}
\end{table*}

In this section, we first detail the experimental settings and the compared methods; then the main results and ablation study are described; finally, we report the human evaluation results and additional analysis based on some cases. Our experiments are conducted on \textit{Restoration-200K} \cite{pan2019improving} \textcolor{black}{and \textit{CANARD} \cite{elgohary-etal-2019-unpack}}.  The  statistics of the datasets  are shown in Table \ref{rstat}.

\subsection{Experiment Settings}\label{setting}
We initialize SARG with RoBERTa-wwm-ext \cite{chinese-bert-wwm} for \textit{Restoration-200k}  and bert-base-uncased \cite{devlin2018bert} for \textit{CANARD}, the hidden size is set to 768, the number of attention heads to 12, the number of attention layers to 12. Adam optimizer is utilized, the loss of tagger weighted to $\alpha = 3$, coverage loss weighted to $\lambda = 1$ and  the initial learning rate is 5e-5. The above hyperparameters are all tuned on the standard validation data.

The according automatic evaluation metrics are utilized as in previus works \cite{pan2019improving, elgohary-etal-2019-unpack}, which contain BLEU, ROUGE, and restoration score. 

\subsection{Compared Methods}
We compare the performance of our proposed SARG with the following methods:
\begin{itemize}
	\item \textbf{CopyNet}: in this baseline, LSTM-based Seq2Seq model with attention and a copy mechanism is employed.
	
	\item \textbf{PAC}\ \cite{pan2019improving}:\ this model restores the incomplete utterance in a cascade way: firstly, select the remained words by fintuning BERT, and then roughly concatenate the selected words, history, original utterance and feed them into a standard pointer-generator network.
	
	\item \textbf{T-Ptr-$\lambda$} \footnote{We re-implement the transformer-based method and evaluate on the same blind test set for the fair comparison.}\ \cite{su2019improving}:\ this model solves such restoration task in an end-to-end way. It employs six layers of transformer blocks as encoder and another six layers of transformer blocks as pointer decoder. Moreover, to emphasize the difference between history and utterance, it takes two individual channels in the encoder-decoder attention.
	
	\item \textbf{Seq2Seq-Uni}:\ we construct this baseline by employing the unified transformer blocks \cite{dong2019unified} as the backbone of Seq2Seq, so that we can load the pretrained transformers easily. 
	
\end{itemize}

\subsection{Main Results}

The main results on \textit{Restoration-200k} are as shown in Table \ref{200k}. Focusing on the automatic metrics, we observe that SARG achieves the best results on 6 of 7 automatic metrics. The superiority of SARG is as expected, on one hand, the words of original utterance can be easily kept or deleted by sequence labeling, on the other hand, the rest of words can be easily copied from history or generated from vocabulary. And we also find PAC is 1.2 higher than SARG on restoration $f_1$ score but 3.0 and 6.1 lower on $f_2$ and $f_3$ separately. In fact, $f_1$ pays more attention to those tokens restored from history than others from the original utterance. In other words, though PAC can recall appropriate restored tokens from history, it may not place these restored tokens in their right positions well. We also exemplify such problem in the case study. Additionally, we compare the results of the beam-search with those of the greedy-earch, which we find
that the beam-search brings pretty significant improvements on those complete autoregressive models, but less obvious on our model. It means that, SARG is less dependent on the beam-search and can be more time-efficient in the inference phase.

\begin{table}[!htb]
	\centering
	\begin{tabular}{lcc}
		\hline
		& Dev  & Test \\ \hline
		CopyNet & 51.37 & 49.67 \\ 
		T-Ptr-$\lambda$ & 46.26 &  45.37 \\ 
		Seq2Seq-Uni &  52.71 & 45.31 \\ 
		SARG & \textbf{56.93} & \textbf{54.80} \\
		Human Rewrites & \multicolumn{2}{c}{59.92} \\
		\hline
		
	\end{tabular}
	\caption{The main results on \textit{CANARD} of our method and other SOTA methods. Table shows the BLEU  scores of the listed models on development and test data.}\label{canard}
\end{table}
Table \ref{canard} shows the main results on \textit{CANARD} dataset. As can be seen, SARG  achieve the best BLEU score\footnote{We use multi-bleu-detok.perl \cite{sennrich-etal-2017-nematus} as in \cite{elgohary-etal-2019-unpack}} on the development and test data. It is 5.56 higher than the previous best on the development data and 5.13 higher on the test data. Moreover, we also find that the result of our method is far from the level of human rewrites, which means that there is still a large room for the improvement of existing rewriting methods.

\begin{table}[!htb]
	\centering
	\begin{tabular}{lc}
		\hline
		& Inference Time \\ \hline
		T-Ptr-$\lambda$ (n\_beam=1)  & 522 s \\
		T-Ptr-$\lambda$ (n\_beam=5) & 602 s \\
		Seq2Seq-Uni (n\_beam=1)  & 321 s  \\
		Seq2Seq-Uni (n\_beam=5)  & 467 s \\
		SARG (n\_beam=1) & 50 s\\
		SARG (n\_beam=5)	 & 70 s \\
		\hline
		
	\end{tabular}
	\caption{The inference time on \textit{Restoration-200k}, which is evaluated on the same blind test set (5104 examples) with one Nvidia Tesla P40. We do not consider the inference speed of PAC, because the cascade way takes lower efficiency than other end-to-end methods} \label{infertime}
\end{table}

Through the Table \ref{infertime}, we can observe that, compared to those complete autoregressive methods, our semi autoregressive model takes less time for inference. SARG is near 10x times as fast as T-Ptr-$\lambda$ and 6x times as fast as Seq2Seq-Uni. Beam-search increases the burden on inference. It needs more time and more memory for maintaining the candidate beams. Generally, the incomplete utterance restoration is required to be time-efficient as the intermediate subtask of multi-turn dialogue task, and it is unpractical to maintain plenty of beams in decoding. Therefore, our SARG may be a suitable choice with the less dependent on the beam-search.

\subsection{Ablation Study}

\begin{CJK}{UTF8}{gbsn}
	\newcommand{\hanzi}{\fontsize{8.5pt}{\baselineskip}\selectfont}
	\newcommand{\tabincell}[2]{\begin{tabular}{@{}#1@{}}#2\end{tabular}}
	\begin{table*}[!htb]
		\centering
		\scalebox{0.8}{
			\begin{tabular}{|c|ccc|}
				\hline
				& \small Example 1 & \small Example 2 & \small Example 3 \\ \hline
				\small $A_1$& \tabincell{c}{\hanzi{男185} \\ \small Male, 185} & \tabincell{c}{\hanzi{天蝎座有喜欢的吗}\\ \small Does anyone like Scorpio} & \tabincell{c}{\hanzi{比尔吉沃特有互送皮肤的吗} \\ \small Who want to exchange skin in Bilgewater} \\
				\small $B_1$& \tabincell{c}{\hanzi{老乡你帖子要沉了}\\ \small Bro, your post is totally ignored} & \tabincell{c}{\hanzi{天蝎座的小伙子你喜欢么} \\ \small Do you like Scorpio boy} & \tabincell{c}{\hanzi{直接买个蘑菇的不就行了} \\ \small Just buy the Timo's directly} \\ 
				\small $A_2$& \tabincell{c}{\hanzi{这不还有你么哈哈} \\ \small You're still here haha} & \tabincell{c}{\hanzi{喜欢啊} \\ \small Yes} & \tabincell{c}{\hanzi{抽奖比较欢乐} \\ \small Lucky draw is interesting} \\ 
				\small $B_2$& \tabincell{c}{\hanzi{你雇我给你吆喝啊} \\ \small You can hire me to cry out for you} & \tabincell{c}{\hanzi{可是我不信星座诶} \\ \small I don't believe in constellations} & \tabincell{c}{\hanzi{我一般都是从我同学的号挨个送我} \\ \small I usually send myself from classmate's account} \\ 
				\small $A_3$& \tabincell{c}{\hanzi{哈哈好} \\ \small OK} & \tabincell{c}{\hanzi{这东西只是娱乐罢} \\ \small It is just entertainment} & \tabincell{c}{\hanzi{看来我只能拿朋友的号送自己了} \\ \small I'll have to send myself from friends' account} \\ \hline
				\small Reference& \tabincell{c}{\hanzi{哈哈\textcolor{red}{给我吆喝}好} \\ \small OK, \textcolor{red}{cry out for me}} & \tabincell{c}{\hanzi{\textcolor{red}{星座}这东西只是娱乐罢} \\ \small \textcolor{red}{Constellation} is just entertainment}  & \tabincell{c}{\hanzi{看来我只能拿朋友的号送\textcolor{red}{皮肤}自己了} \\ \small I'll have to send myself \textcolor{red}{skin} from friends' account} \\ \hline
				\small SARG& \tabincell{c}{\hanzi{哈哈好\textcolor{blue}{雇你给我吆喝}} \\ \small OK, \textcolor{blue}{hire you to cry out for me}} & \tabincell{c}{\hanzi{\textcolor{blue}{星座}这东西只是娱乐罢} \\ \small \textcolor{blue}{Constellation} is just entertainment} & \tabincell{c}{\hanzi{看来我只能拿朋友的号送自己\textcolor{blue}{皮肤}了} \\ \small I'll have to send myself \textcolor{blue}{skin} from friends' account} \\ 
				\small PAC & \tabincell{c}{\hanzi{哈哈\textcolor{blue}{吆喝}好} \\ \small OK, \textcolor{blue}{cry out}} & \tabincell{c}{\hanzi{这东西只是娱乐\textcolor{blue}{星座}罢} \\ \small It is just entertainment \textcolor{blue}{constellation}} & \tabincell{c}{\hanzi{看来我只能拿朋友的号送自己了} \\ \small I'll have to send myself from friends' account} \\ 
				\small T-Ptr-$\lambda$& \tabincell{c}{\hanzi{哈哈好\textcolor{blue}{吆喝}} \\ \small OK, \textcolor{blue}{cry out}} & \tabincell{c}{\hanzi{\textcolor{blue}{不信星座}这东西只是娱乐罢} \\ \small \textcolor{blue}{Not believing constellation} is just entertainment} & \tabincell{c}{\hanzi{看来我只能拿朋友的号送自己了} \\ \small I'll have to send myself from friends' account} \\ 
				\small Seq2Seq-Uni& \tabincell{c}{\hanzi{哈哈好} \\ \small OK} & \tabincell{c}{\hanzi{\textcolor{blue}{不信星座}这东西只是娱乐罢} \\ \small \textcolor{blue}{Not believing constellation} is just entertainment} & \tabincell{c}{\hanzi{看来我只能拿朋友的号送自己了} \\ \small I'll have to send myself from friends' account} \\ \hline
		\end{tabular}}
		\caption{Examples for incomplete utterance restoration. $A_1$ to $B_2$ is the history of conversation, $A_3$ is the original utterance.} \label{caseexample}
	\end{table*}
\end{CJK}

In this subsection, we conduct a series of ablation studies to evaluate the effectiveness of different components in our proposed SARG, which includes pretrained weights (WEIGHT), copy mechanism (COPY), and generation from vocabulary (GEN), and the results are shown in Table \ref{ablation}. 

As can be seen, GEN plays the least important role in our model. By contrast, the absence of COPY or WEIGHT may raise a substantial lack of performance. Following our previous experimental setting, the above two variant models both can not converge well. In fact, the model without COPY only selects words from the pre-defined overall vocabulary, and the decoder is more difficult to be trained well. \textcolor{black}{Furthermore, without the WEIGHT, the model needs to update the overall weights from scratch, which incorporate the 768-dimensional embedding table and 12 transformer layers. Therefore, it is a considerable burden for the optimization, where the limited corpus is provided.} 

And we also compare the output of tagger among the above listed models. An observation is that the tagger without WEIGHT is conservative on predicting the \texttt{CHANGE} operations; by contrast, the decoder without WEIGHT is less affected and has normal-appearing. Therefore, in some cases, even though the decoder produces the right restored words, the model still can not output the correct answers because the tagger does not produce the corresponding \texttt{CHANGE} operations.

\subsection{Human Evaluation}

\begin{table}[!htb]
	\centering
	\begin{tabular}{ccc}
		\hline
		& Quality  & Fluency \\ \hline
		SARG & \textbf{2.70} & 2.85 \\
		PAC & 2.67 & 2.83 \\ 
		T-Ptr-$\lambda$& 2.58 & 2.80 \\ 
		Seq2Seq-Uni& 2.65 & \textbf{2.87} \\ \hline
	\end{tabular}
	\caption{Human evaluation of the restoration quality and language fluency on \textit{Restoration-200k}. Both quality and fluency score adopt a 3-point scale.}\label{human}
\end{table}

In the phase of human evaluation, we employ three experienced workers to score the restoration quality and sentence fluency separately on 200 randomly selected samples. \textcolor{black}{Specifically, each sample is scored by the three workers in turn, and the final quality or fluency scores are calculated by averaging the annotated results}.

As shown in Table \ref{human}, SARG obtains the highest score in restoration quality among the compared methods, which is consistent with the results of automatic evaluation. However, in the aspect of fluency score, Seq2Seq-Uni achieves the best performance. Seq2Seq-Uni takes a way of complete autoregression and benefits from the pretrained weights, which can complete the causal language modeling well. 

\subsection{Case Study}\label{casestudy}

In this subsection, we observe the prediction results among different models, and then select several representative examples to illustrate the superiority of our proposed model as Table \ref{caseexample} shows.

As can be seen in Example 1, the first three models can restore the action ``\textit{cry out}", and only SARG can restore the predicate ``\textit{hire you}", which is important to understand the direction of the action. 

In Example 2, all four models restore the keyword ``\textit{constellation}" correctly. However, for T-Ptr-$\lambda$ and Seq2Seq-Uni, undesired words ``\textit{not believing}" are also restored, which changes the intention of utterance. In PAC, we can find the keyword ``\textit{constellation}" is placed in a wrong position, which leads to the difficulty in understanding. Moreover, for the restoration scores, the wrong position problem has no effect on $f_1$ but is negative for $f_2$ and $f_3$. That is a possible reason, compared with SARG, PAC has higher $f_1$ but lower $f_2$ and $f_3$ in the automatic evaluation.

Finally Example 3 demonstrates the ability of SARG to restore utterance from distant context. Specifically, the keyword ``\textit{skin}" appears in $A_1$, and the model is required to restore it after three utterances.

\section{Conclusion}
\label{conclusion}
In this paper, we propose a novel semi autoregressive generator for multi-turn incomplete utterance restoration. The proposed model takes in the high efficiency of inference time from sequence labeling and the flexibility of generation from autoregressive modeling. Experimental results on two benchmarks demonstrate that the proposed model is significantly superior to other state-of-the-art methods and an appropriate model of utterance restoration for boosting the multi-turn dialogue system.

 \section{Acknowledgments}
We  thank Hongbo Zhang,  Xiaolei Qin, Fuxiao Zhang and all the anonymous reviewers for their valuable comments. 
\bibliographystyle{aaai21}
\bibliography{paper2}

\begin{thebibliography}{42}
\providecommand{\natexlab}[1]{#1}
\providecommand{\url}[1]{\texttt{#1}}
\providecommand{\urlprefix}{URL }
\expandafter\ifx\csname urlstyle\endcsname\relax
  \providecommand{\doi}[1]{doi:\discretionary{}{}{}#1}\else
  \providecommand{\doi}{doi:\discretionary{}{}{}\begingroup
  \urlstyle{rm}\Url}\fi

\bibitem[{Adiwardana et~al.(2020)Adiwardana, Luong, So, Hall, Fiedel,
  Thoppilan, Yang, Kulshreshtha, Nemade, Lu et~al.}]{adiwardana2020towards}
Adiwardana, D.; Luong, M.-T.; So, D.~R.; Hall, J.; Fiedel, N.; Thoppilan, R.;
  Yang, Z.; Kulshreshtha, A.; Nemade, G.; Lu, Y.; et~al. 2020.
\newblock Towards a human-like open-domain chatbot.
\newblock \emph{arXiv preprint arXiv:2001.09977} .

\bibitem[{Athreya, Ngonga~Ngomo, and Usbeck(2018)}]{athreya2018enhancing}
Athreya, R.~G.; Ngonga~Ngomo, A.-C.; and Usbeck, R. 2018.
\newblock Enhancing Community Interactions with Data-Driven Chatbots--The
  DBpedia Chatbot.
\newblock In \emph{Companion Proceedings of the The Web Conference 2018},
  143--146.

\bibitem[{Ba, Kiros, and Hinton(2016)}]{ba2016layer}
Ba, J.~L.; Kiros, J.~R.; and Hinton, G.~E. 2016.
\newblock Layer normalization.
\newblock \emph{arXiv preprint arXiv:1607.06450} .

\bibitem[{Cao et~al.(2018)Cao, Li, Li, and Wei}]{cao2018retrieve}
Cao, Z.; Li, W.; Li, S.; and Wei, F. 2018.
\newblock Retrieve, rerank and rewrite: Soft template based neural
  summarization.
\newblock In \emph{Proceedings of the 56th Annual Meeting of the Association
  for Computational Linguistics (Volume 1: Long Papers)}, 152--161.

\bibitem[{Chen and Bansal(2018)}]{chen2018fast}
Chen, Y.-C.; and Bansal, M. 2018.
\newblock Fast abstractive summarization with reinforce-selected sentence
  rewriting.
\newblock \emph{arXiv preprint arXiv:1805.11080} .

\bibitem[{Chollampatt and Ng(2018)}]{chollampatt2018neural}
Chollampatt, S.; and Ng, H.~T. 2018.
\newblock Neural quality estimation of grammatical error correction.
\newblock In \emph{Proceedings of the 2018 Conference on Empirical Methods in
  Natural Language Processing}, 2528--2539.

\bibitem[{Cui et~al.(2019)Cui, Che, Liu, Qin, Yang, Wang, and
  Hu}]{chinese-bert-wwm}
Cui, Y.; Che, W.; Liu, T.; Qin, B.; Yang, Z.; Wang, S.; and Hu, G. 2019.
\newblock Pre-Training with Whole Word Masking for Chinese BERT.
\newblock \emph{arXiv preprint arXiv:1906.08101} .

\bibitem[{Devlin et~al.(2018)Devlin, Chang, Lee, and
  Toutanova}]{devlin2018bert}
Devlin, J.; Chang, M.-W.; Lee, K.; and Toutanova, K. 2018.
\newblock Bert: Pre-training of deep bidirectional transformers for language
  understanding.
\newblock \emph{arXiv preprint arXiv:1810.04805} .

\bibitem[{Dong et~al.(2019)Dong, Yang, Wang, Wei, Liu, Wang, Gao, Zhou, and
  Hon}]{dong2019unified}
Dong, L.; Yang, N.; Wang, W.; Wei, F.; Liu, X.; Wang, Y.; Gao, J.; Zhou, M.;
  and Hon, H.-W. 2019.
\newblock Unified language model pre-training for natural language
  understanding and generation.
\newblock In \emph{Advances in Neural Information Processing Systems},
  13063--13075.

\bibitem[{Elgohary, Peskov, and Boyd-Graber(2019)}]{elgohary-etal-2019-unpack}
Elgohary, A.; Peskov, D.; and Boyd-Graber, J. 2019.
\newblock Can You Unpack That? Learning to Rewrite Questions-in-Context.
\newblock In \emph{Proceedings of the 2019 Conference on Empirical Methods in
  Natural Language Processing and the 9th International Joint Conference on
  Natural Language Processing (EMNLP-IJCNLP)}, 5918--5924. Hong Kong, China:
  Association for Computational Linguistics.
\newblock \doi{10.18653/v1/D19-1605}.
\newblock \urlprefix\url{https://www.aclweb.org/anthology/D19-1605}.

\bibitem[{Ge, Wei, and Zhou(2018)}]{ge2018fluency}
Ge, T.; Wei, F.; and Zhou, M. 2018.
\newblock Fluency boost learning and inference for neural grammatical error
  correction.
\newblock In \emph{Proceedings of the 56th Annual Meeting of the Association
  for Computational Linguistics (Volume 1: Long Papers)}, 1055--1065.

\bibitem[{Gong et~al.(2019)Gong, Kong, Zhang, Tan, Zhang, and
  Shao}]{gong2019customer}
Gong, X.; Kong, X.; Zhang, Z.; Tan, L.; Zhang, Z.; and Shao, B. 2019.
\newblock Customer Service Automatic Answering System Based on Natural Language
  Processing.
\newblock In \emph{Proceedings of the 2019 International Symposium on Signal
  Processing Systems}, 115--120.

\bibitem[{Hewitt and Beaver(2020)}]{hewitt2020case}
Hewitt, T.; and Beaver, I. 2020.
\newblock A Case Study of User Communication Styles with Customer Service
  Agents versus Intelligent Virtual Agents.
\newblock In \emph{Proceedings of the 21th Annual Meeting of the Special
  Interest Group on Discourse and Dialogue}, 79--85.

\bibitem[{Hochreiter and Schmidhuber(1997)}]{hochreiter1997long}
Hochreiter, S.; and Schmidhuber, J. 1997.
\newblock Long short-term memory.
\newblock \emph{Neural computation} 9(8): 1735--1780.

\bibitem[{Huang, Zhu, and Gao(2020)}]{huang2020challenges}
Huang, M.; Zhu, X.; and Gao, J. 2020.
\newblock Challenges in building intelligent open-domain dialog systems.
\newblock \emph{ACM Transactions on Information Systems (TOIS)} 38(3): 1--32.

\bibitem[{Ji, Lu, and Li(2014)}]{ji2014information}
Ji, Z.; Lu, Z.; and Li, H. 2014.
\newblock An information retrieval approach to short text conversation.
\newblock \emph{arXiv preprint arXiv:1408.6988} .

\bibitem[{Lebanoff et~al.(2019)Lebanoff, Muchovej, Dernoncourt, Kim, Kim,
  Chang, and Liu}]{lebanoff2019analyzing}
Lebanoff, L.; Muchovej, J.; Dernoncourt, F.; Kim, D.~S.; Kim, S.; Chang, W.;
  and Liu, F. 2019.
\newblock Analyzing sentence fusion in abstractive summarization.
\newblock \emph{arXiv preprint arXiv:1910.00203} .

\bibitem[{Li(2020)}]{li2020empirical}
Li, P. 2020.
\newblock An Empirical Investigation of Pre-Trained Transformer Language Models
  for Open-Domain Dialogue Generation.
\newblock \emph{arXiv preprint arXiv:2003.04195} .

\bibitem[{Lin et~al.(2020)Lin, Zhang, Liu, and Ma}]{lin2020hierarchical}
Lin, F.; Zhang, C.; Liu, S.; and Ma, H. 2020.
\newblock A Hierarchical Structured Multi-Head Attention Network for Multi-Turn
  Response Generation.
\newblock \emph{IEEE Access} 8: 46802--46810.

\bibitem[{Malmi et~al.(2019)Malmi, Krause, Rothe, Mirylenka, and
  Severyn}]{malmi2019encode}
Malmi, E.; Krause, S.; Rothe, S.; Mirylenka, D.; and Severyn, A. 2019.
\newblock Encode, tag, realize: High-precision text editing.
\newblock \emph{arXiv preprint arXiv:1909.01187} .

\bibitem[{Ng et~al.(2014)Ng, Wu, Briscoe, Hadiwinoto, Susanto, and
  Bryant}]{ng2014conll}
Ng, H.~T.; Wu, S.~M.; Briscoe, T.; Hadiwinoto, C.; Susanto, R.~H.; and Bryant,
  C. 2014.
\newblock The CoNLL-2014 shared task on grammatical error correction.
\newblock In \emph{Proceedings of the Eighteenth Conference on Computational
  Natural Language Learning: Shared Task}, 1--14.

\bibitem[{Pan et~al.(2019)Pan, Bai, Wang, Zhou, and Liu}]{pan2019improving}
Pan, Z.; Bai, K.; Wang, Y.; Zhou, L.; and Liu, X. 2019.
\newblock Improving open-domain dialogue systems via multi-turn incomplete
  utterance restoration.
\newblock In \emph{Proceedings of the 2019 Conference on Empirical Methods in
  Natural Language Processing and the 9th International Joint Conference on
  Natural Language Processing (EMNLP-IJCNLP)}, 1824--1833.

\bibitem[{Radford et~al.(2018)Radford, Narasimhan, Salimans, and
  Sutskever}]{radford2018improving}
Radford, A.; Narasimhan, K.; Salimans, T.; and Sutskever, I. 2018.
\newblock Improving language understanding by generative pre-training.

\bibitem[{Ritter, Cherry, and Dolan(2011)}]{ritter2011data}
Ritter, A.; Cherry, C.; and Dolan, W.~B. 2011.
\newblock Data-driven response generation in social media.
\newblock In \emph{Proceedings of the 2011 Conference on Empirical Methods in
  Natural Language Processing}, 583--593.

\bibitem[{See, Liu, and Manning(2017)}]{see2017get}
See, A.; Liu, P.~J.; and Manning, C.~D. 2017.
\newblock Get to the point: Summarization with pointer-generator networks.
\newblock \emph{arXiv preprint arXiv:1704.04368} .

\bibitem[{Sennrich et~al.(2017)Sennrich, Firat, Cho, Birch, Haddow, Hitschler,
  Junczys-Dowmunt, L{\"a}ubli, Miceli~Barone, Mokry, and
  N{\u{a}}dejde}]{sennrich-etal-2017-nematus}
Sennrich, R.; Firat, O.; Cho, K.; Birch, A.; Haddow, B.; Hitschler, J.;
  Junczys-Dowmunt, M.; L{\"a}ubli, S.; Miceli~Barone, A.~V.; Mokry, J.; and
  N{\u{a}}dejde, M. 2017.
\newblock {N}ematus: a Toolkit for Neural Machine Translation.
\newblock In \emph{Proceedings of the Software Demonstrations of the 15th
  Conference of the {E}uropean Chapter of the Association for Computational
  Linguistics}, 65--68. Valencia, Spain: Association for Computational
  Linguistics.
\newblock \urlprefix\url{https://www.aclweb.org/anthology/E17-3017}.

\bibitem[{Serban et~al.(2017)Serban, Klinger, Tesauro, Talamadupula, Zhou,
  Bengio, and Courville}]{serban2017multiresolution}
Serban, I.~V.; Klinger, T.; Tesauro, G.; Talamadupula, K.; Zhou, B.; Bengio,
  Y.; and Courville, A. 2017.
\newblock Multiresolution recurrent neural networks: An application to dialogue
  response generation.
\newblock In \emph{Thirty-First AAAI Conference on Artificial Intelligence}.

\bibitem[{Serban et~al.(2016)Serban, Sordoni, Bengio, Courville, and
  Pineau}]{serban2016building}
Serban, I.~V.; Sordoni, A.; Bengio, Y.; Courville, A.; and Pineau, J. 2016.
\newblock Building end-to-end dialogue systems using generative hierarchical
  neural network models.
\newblock In \emph{Thirtieth AAAI Conference on Artificial Intelligence}.

\bibitem[{Su et~al.(2019)Su, Shen, Zhang, Sun, Hu, Niu, and
  Zhou}]{su2019improving}
Su, H.; Shen, X.; Zhang, R.; Sun, F.; Hu, P.; Niu, C.; and Zhou, J. 2019.
\newblock Improving multi-turn dialogue modelling with utterance ReWriter.
\newblock \emph{arXiv preprint arXiv:1906.07004} .

\bibitem[{Thadani and McKeown(2013)}]{thadani2013supervised}
Thadani, K.; and McKeown, K. 2013.
\newblock Supervised sentence fusion with single-stage inference.
\newblock In \emph{Proceedings of the Sixth International Joint Conference on
  Natural Language Processing}, 1410--1418.

\bibitem[{Vaswani et~al.(2017)Vaswani, Shazeer, Parmar, Uszkoreit, Jones,
  Gomez, Kaiser, and Polosukhin}]{vaswani2017attention}
Vaswani, A.; Shazeer, N.; Parmar, N.; Uszkoreit, J.; Jones, L.; Gomez, A.~N.;
  Kaiser, {\L}.; and Polosukhin, I. 2017.
\newblock Attention is all you need.
\newblock In \emph{Advances in neural information processing systems},
  5998--6008.

\bibitem[{Wu, Wang, and Xue(2016)}]{wu2016ranking}
Wu, B.; Wang, B.; and Xue, H. 2016.
\newblock Ranking responses oriented to conversational relevance in chat-bots.
\newblock In \emph{Proceedings of COLING 2016, the 26th International
  Conference on Computational Linguistics: Technical Papers}, 652--662.

\bibitem[{Wu et~al.(2016)Wu, Wu, Xing, Zhou, and Li}]{wu2016sequential}
Wu, Y.; Wu, W.; Xing, C.; Zhou, M.; and Li, Z. 2016.
\newblock Sequential matching network: A new architecture for multi-turn
  response selection in retrieval-based chatbots.
\newblock \emph{arXiv preprint arXiv:1612.01627} .

\bibitem[{Wubben, Krahmer, and van~den Bosch(2012)}]{wubben2012sentence}
Wubben, S.; Krahmer, E.; and van~den Bosch, A. 2012.
\newblock Sentence simplification by monolingual machine translation .

\bibitem[{Xing et~al.(2016)Xing, Wu, Wu, Liu, Huang, Zhou, and
  Ma}]{xing2016topic}
Xing, C.; Wu, W.; Wu, Y.; Liu, J.; Huang, Y.; Zhou, M.; and Ma, W.-Y. 2016.
\newblock Topic augmented neural response generation with a joint attention
  mechanism.
\newblock \emph{arXiv preprint arXiv:1606.08340} 2(2).

\bibitem[{Yan, Song, and Wu(2016)}]{yan2016learning}
Yan, R.; Song, Y.; and Wu, H. 2016.
\newblock Learning to respond with deep neural networks for retrieval-based
  human-computer conversation system.
\newblock In \emph{Proceedings of the 39th International ACM SIGIR conference
  on Research and Development in Information Retrieval}, 55--64.

\bibitem[{Zhang and Lapata(2017)}]{zhang2017sentence}
Zhang, X.; and Lapata, M. 2017.
\newblock Sentence Simplification with Deep Reinforcement Learning.
\newblock In \emph{Proceedings of the 2017 Conference on Empirical Methods in
  Natural Language Processing}, 584--594.

\bibitem[{Zhang et~al.(2018)Zhang, Li, Zhu, Zhao, and Liu}]{zhang2018modeling}
Zhang, Z.; Li, J.; Zhu, P.; Zhao, H.; and Liu, G. 2018.
\newblock Modeling multi-turn conversation with deep utterance aggregation.
\newblock \emph{arXiv preprint arXiv:1806.09102} .

\bibitem[{Zhao et~al.(2019)Zhao, Wang, Shen, Jia, and Liu}]{zhao2019improving}
Zhao, W.; Wang, L.; Shen, K.; Jia, R.; and Liu, J. 2019.
\newblock Improving grammatical error correction via pre-training a
  copy-augmented architecture with unlabeled data.
\newblock \emph{arXiv preprint arXiv:1903.00138} .

\bibitem[{Zhao, Xu, and Wu(2020)}]{zhao2020learning}
Zhao, Y.; Xu, C.; and Wu, W. 2020.
\newblock Learning a Simple and Effective Model for Multi-turn Response
  Generation with Auxiliary Tasks.
\newblock \emph{arXiv preprint arXiv:2004.01972} .

\bibitem[{Zhou et~al.(2016)Zhou, Dong, Wu, Zhao, Yu, Tian, Liu, and
  Yan}]{zhou2016multi}
Zhou, X.; Dong, D.; Wu, H.; Zhao, S.; Yu, D.; Tian, H.; Liu, X.; and Yan, R.
  2016.
\newblock Multi-view response selection for human-computer conversation.
\newblock In \emph{Proceedings of the 2016 Conference on Empirical Methods in
  Natural Language Processing}, 372--381.

\bibitem[{Zhou et~al.(2018)Zhou, Li, Dong, Liu, Chen, Zhao, Yu, and
  Wu}]{zhou2018multi}
Zhou, X.; Li, L.; Dong, D.; Liu, Y.; Chen, Y.; Zhao, W.~X.; Yu, D.; and Wu, H.
  2018.
\newblock Multi-turn response selection for chatbots with deep attention
  matching network.
\newblock In \emph{Proceedings of the 56th Annual Meeting of the Association
  for Computational Linguistics (Volume 1: Long Papers)}, 1118--1127.

\end{thebibliography}

\end{document}